\documentclass[10pt,twocolumn,letterpaper]{article}

\usepackage{cvpr}              % To produce the CAMERA-READY version
\definecolor{cvprblue}{rgb}{0.21,0.49,0.74}
\usepackage[pagebackref,breaklinks,colorlinks,allcolors=cvprblue]{hyperref}
\usepackage{tabularx}
\usepackage{booktabs}
\usepackage{colortbl}
\usepackage{graphicx}
\usepackage{amsmath} 
\usepackage{indentfirst}
\usepackage{algorithm}

\usepackage{algpseudocode}

\usepackage{multirow}

\definecolor{pink}{HTML}{C0E4FC}
\def\paperID{7347} % *** Enter the Paper ID here
\def\confName{CVPR}
\def\confYear{2025}

\title{PAR: Prompt-Aware Token Reduction Method  \\
            for Efficient Large Multimodal Models}

\author{
    Yingen Liu, Fan Wu, Ruihui Li, Zhuo Tang, Kenli Li \\
     College of Computer Science and Electronic Engineering, Hunan University\\
     \texttt{\{liuyingen, wufan, liruihui, ztang, lkl\}@hnu.edu.cn}\\
     }

\begin{document}
\maketitle
\begin{abstract}
Multimodal large language models (MLLMs) demonstrate strong performance across visual tasks, but their efficiency is hindered by significant computational and memory demands from processing long contexts in multimodal inputs. 
To address this, we introduce PAR (Prompt-Aware Token Reduction), a novel and plug-and-play approach that reduces visual tokens efficiently without compromising model performance. Unlike previous methods that rely heavily on attention mechanisms and overlooking cross-modal interactions , we uses a prompt-aware strategy to adpative identify and cluster essential visual tokens.
PAR categorizes visual context redundancy into two types: external and internal. External redundancy is minimized through semantic retrieval, while internal redundancy is addressed using a token routing mechanism. This method substantially reduces computational load without requiring additional training or complex architectural modifications. 
\textbf{Experimental results demonstrate that across various visual question answering tasks, PAR reduces FLOPs by 83\% with a compression ratio of 89\%, while retaining 97\% of baseline accuracy.}
The adaptive design of PAR achieves a 2x token reduction ratio compared to prior approaches, enabling a better balance between performance and efficiency.

\end{abstract}    

\section{Introduction}

Thanks to advanced model architectures and extensive training data, large language models(LLMs) have achieved exceptional performance across various application domains. Traditionally, these models have focused on textual inputs, excelling in tasks like natural language understanding and generation within prominent NLP fields\cite{touvron2023} \cite{chowdhery2023palm} \cite{ouyang2022training}. However, real-world data includes not only text but also diverse modalities such as images, audio recordings, and point clouds. 

Researchers are now exploring the extension of these impressive capabilities into multimodal domains\cite{jin2024}\cite{song2023}, leading to the development of multimodal large language models (MLLMs)  such as GPT-4\cite{openai2024}, Gemini\cite{geminiteam2024}, LLaVA\cite{liu2023}, and MiniGPT-4\cite{zhu2023}. These models avoid the high computational costs associated with training from scratch by leveraging pre-training knowledge specific to each modality. Moreover, by establishing of strong representational mappings and alignments with modality-specific models, these multimodal models can efficiently process inputs across multiple modalities, significantly broadening their potential application areas.

\begin{figure}[t]
    \hfill % 将图像推向右边
     \hspace{-1em} % 调整左侧位置
     \includegraphics[width=1.0\linewidth]{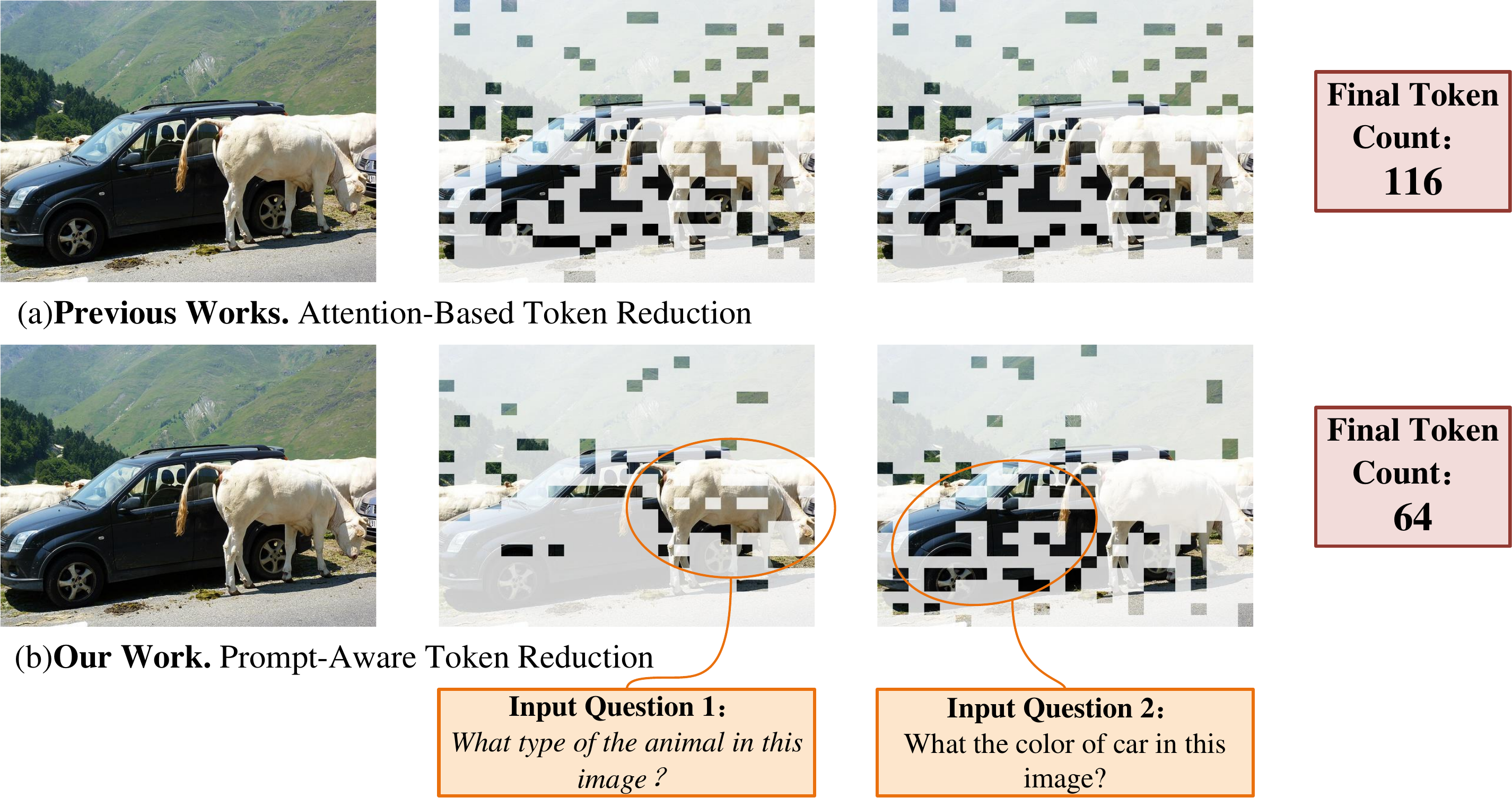} % 适当调整宽度
      % \vspace{-1em} % 减小标题与图片之间的距离
    \caption{
Previous works relying on the attention mechanism, focus on global  visual tokens and cause unnecessary redundancy.
In contrast, our method is guided by prompts and focuses more effectively on the task-relevant visual tokens. \textbf{Our approach achieves a token reduction ratio about 2x of previous methods.}}
\label{fig:enter-label}
\end{figure}

% Despite their potential, MLMMs face significant challenges during deployment and inference, primarily related to computational resources and efficiency. MLMMs rely on visual encoders to transform image input into large parts of visual tokens, which are then concatenated with user and system prompts to form a longer sequence for generation. The length of this sequence leads to increased storage and computational demands. \textit{How to reducing the number of visual tokens and without significantly impacting generation performance is a critical challenge that must be addressed.}
Despite their potential, MLLMs face significant computational and memory challenges, particularly in deployment and inference. MLLMs rely on visual encoders to convert image inputs into large sets of visual tokens, concatenated with prompt tokens, forming long input sequences. This increase in sequence length raises storage and computational demands. How to reduce the number of visual tokens without significantly impacting generation performance is a critical challenge that must be addressed.

Traditional token reduction methods\cite{chen2023} \cite{shang2024} \cite{chen2025image} typically rely on attention mechanisms to identify and remove unimportant tokens. 
Attention mechanisms capture global semantic information and measure token importance based on attention scores, but they struggle to adaptively balance efficiency and performance according to specific task requirements.
In tasks requiring strong cross-modal interactions, such as vision-language understanding, reliance on attention mechanisms often introduces excessive redundancy or leads to significant drops in task accuracy.
Prior works also have failed to fully exploit the unique redundancy of visual tokens, depending on conventional pruning methods to optimize MLMMs.
In addition, previous research frequently requires custom network architectures \cite{chu2023} \cite{cao2024madtp} \cite{Haurum_2023_ICCV}, further increasing computational costs.

Fully considering both external and internal redundancy observed in visual tokens, and drawing inspiration from human behavior in visual question-answering tasks, we introduce a simple yet effective training-free token reduction method called PAR (\textbf{P}rompt-\textbf{A}ware Token \textbf{R}eduction). In this approach, we develop a prompt-aware strategy to identify and retrieve important visual tokens within the given context, effectively optimizing the selection of tokens based on their semantic relevance.

Our method proceeds as follows. First, we apply predefined templates to rewrite the user prompt, enhancing its semantic representation in future steps. Next, we use a graph-based clustering algorithm to partition visual tokens into meaningful semantic clusters based on similarity distribution. We then perform prompt-guided semantic retrieval, matching the most prompt-relevant visual tokens to eliminate external redundancy. Finally, a token router simplifies the retained tokens by refining the final selection thus removing internal redundancy.

In the design of the experiments, we fully considered the balance between performance and efficiency. \textit{Our goal was to achieve the optimal trade-off by minimizing the final number of visual tokens used while ensuring a minimal loss in accuracy.}
Experimental results show that PAR achieves an 83\% reduction in FLOPs and an 89\% compression ratio across diverse visual question-answering tasks, all while preserving 97\% of the baseline accuracy.
Notably, in hallucination benchmarks\cite{li2023},  PAR outperforms the original model under specific settings, indicating effective mitigation of hallucination phenomena in MLLMs through reduced external redundancy.

 In summary, our work makes three main contributions:
    \begin{itemize}
        \item We conducted an in-depth analysis of the redundancy present in visual representations, categorizing it into external and internal redundancy. External redundancy refers to representations irrelevant to the task, while internal redundancy represents those that contribute overlapping semantic information.
        \item 
        Inspired by human cognition, we propose \textbf{PAR}, a training-free method for reducing visual tokens. PAR leverages prompt semantics to eliminate external redundancy, while a token router adaptively filters internal redundancies, retaining only the most relevant tokens.
        
        \item Experimental results show that PAR not only surpasses previous methods in accuracy but also achieves a \textbf{2x} token reduction, effectively balancing efficiency and performance.
    \end{itemize}

\section{Related works}
\subsection{Multimodal Large Language Models}

The development of Large Language Models (LLMs) such as GPT-3\cite{brown2020}, LLaMA\cite{touvron2023}, and GPT-4\cite{openai2024} has seen substantial progress in recent years. These advancements have inspired the evolution of Multimodal Large Language Models (MLLMs), which extend the capabilities of LLMs to include images. Notable examples of this progress are LLava\cite{liu2023}, MiniGPT-4\cite{zhu2023}, InstructBLIP\cite{dai2023},Qwen\cite{bai2023},and Gemini\cite{geminiteam2024} .

These MLLMs primarily utilize a visual encoder\cite{radford2021} to process visual input. They then align this visual data with text through linear projection and concatenate the visual information with text tokens for generation by a pre-trained LLM. By integrating data from various modalities, MLLMs enhance contextual understanding, thereby improving the accuracy of information processing and generation.

Despite these advancements, MLLMs face significant computational costs during inference and deployment, highlighting the need for efficient token reduction techniques. Especially when dealing with videos or high-resolution images, processing thousands of tokens becomes necessary.

\subsection{Visual token pruning}

The quadratic complexity of Transformers \cite{vaswani2017} is a significant challenge, especially for MLLMs where image inputs are converted into numerous tokens. This process results in substantial computational costs and limits scalability due to the high memory demands of processing long sequences.

Recent research has sought to improve inference efficiency by pruning visual tokens. LLaVA-PruMerge \cite{shang2024} employs an adaptive visual token reduction strategy that takes advantage of the sparsity in visual encoders, selectively retaining essential tokens and enhancing their informational content through clustering and weighted averaging.  FastV \cite{chen2025image} reduces inference costs by learning adaptive attention patterns in early layers and pruning tokens in later layers. However, these approaches still rely heavily on attention mechanisms and often overlook the cross-modal relevance of tokens.
Due to the inherent properties of the attention mechanism, they struggle to adaptively reduce tokens within a task-aware context, making it difficult to maintain a balance between performance and efficiency.

\section{Observations}

\subsection{The semantic retrieval between modals.}
\label{sec:3.1}

% Humans often exhibit a strong selective focus when performing visual tasks, especially when addressing visual understanding tasks involving text-based questions. They tend to pay more attention to image regions highly relevant to the text, which helps ignore irrelevant areas and reach answers more quickly. Inspired by this behavior, we can leverage prompt to guide MLMMs to perform multimodal tasks more efficiently.

Humans often exhibit selective focus when performing visual tasks, especially when tackling visual understanding tasks that involve text-based questions. They tend to concentrate on image regions that are highly relevant to the accompanying text, helping to ignore irrelevant areas and arrive at answers more efficiently. Inspired by this behavior, we leverage prompts to guide MLMMs, enabling them to perform multimodal tasks with greater efficiency.

Contrastive learning \cite{radford2021} embeds multimodal data in a shared semantic space, providing a solid foundation for identifying relevant semantics across different modalities. 
Rather than relying on attention mechanisms to capture key semantic tokens, we use prompts to guide the retrieval of the most relevant context within the visual input, focusing on visual tokens that are closely related to the task.

\subsection{The redundancy of visual context.} 
\label{sec:3.2}

\textbf{External Redundancy.}Previous research \cite{liu2023deja} \cite{dong2023}has demonstrated that there is substantial external redundancy present within the context of language models. The attention mechanism tends to focus on specific tokens and the latest tokens in the sequence. Tokens that contribute little to the generation process can be considered unimportant redundant tokens. Similarly, visual tokens often concentrate on particular patches during the generation. Large parts of visual tokens are meaningless in generation. The presence of such external redundancy not only augments the computational burden associated with model inference but also has the potential to induce erroneous hallucinations in LLMs.

Previous approaches often rely on attention mechanisms to expel the unimportant tokens but overlook the compactness provided by task-specific prompts, resulting in inefficiencies.
We propose a prompt-aware strategy to focus specifically on visual contexts most relevant to generation tasks, thus effectively eliminating external redundancy.

\textbf{Internal Redundancy.}Unlike text tokens, visual tokens also exhibit a high degree of internal redundancy. Many images may feature similar visual characteristics, such as color, shape, or texture. This similarity necessitates that the model manages a substantial amount of overlapping information when processing images, thereby inducing the issue of redundancy.

Few works address internal redundancy in visual contexts. The main challenge in resolving internal redundancy lies in identifying critical tokens and eliminating repetitive ones. To tackle this, we design a token router based on semantic retrieval ordering and a routing threshold, which preserves distinctive visual tokens and effectively eliminates internal redundancy. This approach ultimately achieves a more optimal balance between performance and efficiency than previous methods.

% We considering that not all image tokens from the visual encoder contribute positively to the final output of the MLMMs. Due to the existence of external redundancy and internal redundancy, a considerable number of visual tokens are likely redundant. The LLM itself possesses sufficient generalization capabilities, suggesting that even a small subset of visual tokens can provide enough semantic information to assist the multimodal large model in the generation process. In summary, redundancy in the visual context can be extraneous for MLLMs and may lead to unnecessary computational expenses .

\section{Methods}

\subsection{Preliminaries}

\begin{figure*}[t]
    \centering
    \includegraphics[width=1\linewidth]{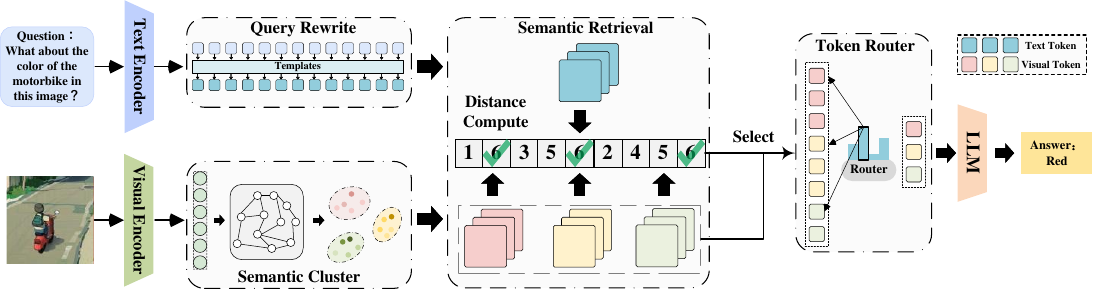}
    \caption{\textbf{The framework of our method.}  Given an input of image and text, PAR processes each modality separately: the text is structured using predefined templates, and the image undergoes semantic clustering. Prompt tokens are then retrieval with visual tokens to select relevant ones, reducing external redundancy. Finally, the token router refines these selections, removing internal redundancy before passing them to the language model (LM) for the answer generation.}
    \label{fig:enter-label}
\end{figure*}

Multimodal large language models (MLLMs) represent a significant advancement in artificial intelligence by integrating visual encoders with pre-trained large language models. This integration allows these models to jointly process and understand diverse modalities, such as images and text, enabling more comprehensive and contextually aware analyses.

For a given image \(I\), the visual encoder \(f_{visual}\) transforms it into a series of token representations \(X_I\):

\begin{equation}
    X_I = f_{visual}(I)
\end{equation}

Here, \(X_I\) denotes a set of visual tokens that capture essential visual features and semantic information from the image. This process allows image content to be encoded in a format compatible with the language model, facilitating the effective integration of visual and textual data.

To ensure that the visual tokens are compatible with the language model's processing framework, a visual projection layer is employed. This layer converts visual tokens \(X_I\) into language embedding tokens \(y_1, y_2, \dots, y_n\). The transformation is performed such that these language embedding tokens share the same dimensional space \(d\) as the word embeddings used by the language model:

\begin{equation}
    \mathbf{y}_j = f_{projection}(\mathbf{X}_I)
\end{equation}

where \(f_{projection}\) denotes the projection function that aligns the dimensionality of the visual tokens with that of the language embeddings.

This alignment ensures that the visual information encoded in \(y_1, y_2, \dots, y_n\) can be processed seamlessly alongside the textual information by the language model. Consequently, the MLMM can effectively integrate and leverage both visual and textual representations, improving its ability to generate accurate and contextually relevant outputs.

\subsection{Semantic Retrieval Improves Multi-Modal Inference}

% Drawing inspiration from human cognitive processes, especially in tasks requiring the integration of visual and textual information, we propose the concept of \textit{semantic retrievaling}. This mirrors the human tendency to focus on image regions that are most relevant to the associated text, allowing for more efficient interpretation and understanding of the multi modal tasks.
To accelerate computation and alleviate the computational burden during inference, an effective and straightforward strategy involves the removal of unimportant tokens. The key to this approach lies in accurately identifying and selecting the important tokens, as this directly influences both the model's inference efficiency and the quality of the final output.

As discussed in  Sec \ref{sec:3.1}. In inference of MLMMs, the output generation is significantly influenced by visual tokens that are closely related to the prompt tokens. We regard these tokens as key tokens to the generation process.

In this paper, we propose using \textbf{semantic retrieval} \cite{borgeaud2022}to determine the important visual tokens.
Specifically, in our approach, the visual vector \( \mathbf{X}_I \in \mathbb{R}^d \) is processed through the visual encoder and serves as candidate, while the text vector \( \mathbf{X}_T \in \mathbb{R}^d \) acts as the query vector. During multimodal inference, the  \( \mathbf{X}_T \) encapsulates the information from the text modality, and \( \mathbf{X}_I \) encodes features from the image modality.

% In retrieval process,  \( \mathbf{X}_T \) is used as query vector and treat \( \mathbf{X}_I \) as candidates, selecting the top-k vectors most similar to the query as the final tokens.
% These vectors coexist in a shared semantic space, enabling cross-modal semantic alignment. To measure the distance between the text and visual representations, we compute a distance score based on a suitable  function \( f \). This function \( f(\mathbf{X}_T, \mathbf{X}_I) \) is chosen to effectively capture the degree of alignment between \( \mathbf{X}_T \) and \( \mathbf{X}_I \), with higher scores indicating stronger alignment.
% Mathematically, this relationship can be formulated as:
In this retrieval process, \( \mathbf{X}_T \) is used as the query vector to select the top-k vectors from \( \mathbf{X}_I \) that are most similar to the query. Both vectors reside in a shared semantic space, facilitating cross-modal alignment. To assess similarity between text and visual representations, we compute a similarity score using a carefully chosen distance function \( f \), where \( f(\mathbf{X}_T, \mathbf{X}_I) \) effectively quantifies the semantic retrieval between \( \mathbf{X}_T \) and each candidate in \( \mathbf{X}_I \), with higher scores indicating greater alignment.

Formally, this retrieval process can be expressed as:
\begin{equation}
    X_{key} = \left\{ X' \mid X' \in \text{Top-}k\left( {f}(X_T, X_I) \right) \right\}
\end{equation}

Here, \(X_{key}\) denotes the most important  visual tokens  that are directly relevant to the prompt.

% MLMMs comes from contrastive learning paradigm, where text and images are mapped to the same semantic space. By calculating vector similarity, the semantic matching degree between different modalities can be effectively assessed. This characteristic allows us to leverage retrieval algorithms to capture the visual tokens generated during the semantic matching process, facilitating a more precise pruning strategy. 

Using semantic retrieval, our approach identifies the most semantically relevant features across modalities, supporting efficient multimodal inference while reducing the computational cost by filtering out less relevant tokens.

% where a higher similarity score indicates a stronger alignment between the text query and visual content.

% This process effectively identifies the most relevant modality information based on semantic matching, providing precise cues for multimodal inference. By filtering out less relevant tokens based on similarity thresholds, the approach also significantly reduces the computational burden, retaining only the most semantically relevant features for further processing.

% This method offers two primary advantages in enhancing the generation process:

% \textbf{(1) Reduction of Computational Load:} By classifying visual tokens into matching and non-matching categories, semantic matching allows for the removal of non-matching tokens. This reduces the computational burden, as the model only processes and generates output based on relevant visual information, thus optimizing performance and efficiency.

% \textbf{(2) Mitigation of Illusion Phenomena:} In MLMMs, the visual domain can often suffer from "illusion phenomena" where irrelevant or misleading visual information affects the model’s output. Semantic matching helps to alleviate this issue by ensuring that the multi-modal model focuses primarily on image regions that are directly related to the textual prompt. This results in more accurate and contextually relevant generation.

\subsection{Token reduction via  hybrid semantic retrieval}
\subsubsection{Pre-Retrieval:Query Rewrite}
% It is not appropriate to use all text tokens as query for retrieval.The multi-modal model employs contrastive learning to embed images and text into a shared semantic space. In future steps, the role of the query prompt is to generate a query vector for text embeddings used in retrieval. However, the original prompts may contain biases or noise, which can lead to certain types of semantic information being inadequately represented in the embedding space, thereby affecting the accuracy of semantic retrievaling.

It is not appropriate to use all prompt tokens as the query for retrieval. Original prompts may contain biases or noise that inadequately represent certain semantic information in the embedding space, thus impacting retrieval accuracy.

To address the issue of information asymmetry between different modalities, we rewrite the query prompts ahead of retrieval. By intentionally controlling the structure and word choices of the query, we ensure that the text embeddings more accurately capture the semantic features relevant to the target image, thus enhancing the performance and retrieval precision.

We simply employ a text rewriting framework based on predefined templates, as illustrated below:

\begin{align}
<Prefix><Core Object><Additional Info>
\end{align}
This framework systematically organizes different forms of text descriptions to ensure that the prompts effectively convey semantic information aligned with visual features. Additionally, key elements can be precisely adjusted based on task requirements to optimize the retrieval performance of the text prompts.

\subsubsection{Hybrid Granularity Retrieval}
The advantage of our method lies in its ability to fully exploit the latent interactions between different modals during multi-modal generating, thereby avoiding inference mistakes that may arise from relying on a single modal. By integrating the retrieval mechanism into multi-modal inference, we achieve more efficient cross-modal information fusion and reasoning, thereby enhancing the overall performance of the model. 
To achieve efficient retrieval, we need to address a pivotal question: \textit{How can we precisely retrieve enough visual tokens that exhibit adequate semantics?}

As discussed in Sec \ref{sec:3.2}, the visual representations exhibit a high degree of internal redundancy. Contrastive learning paradigm \cite{radford2021} does not effectively partition images based on semantics, as each token can only represent information from a single patch, failing to consider the semantic information of neighboring patches. Besides, smaller objects also need to be treated as a basic semantic unit and incorporated into the retrieval process.
An effective retrieval strategy must consider both detailed semantic information and global semantic context.

% Considering above, we thought single-granularity retrieval may face issues of semantic inaccuracy. While fine-grained retrieval can handle more nuanced semantic features, it may overlook the overall contextual information; conversely, coarse-grained retrieval, although capturing the consistency of local semantics, may fall short when addressing detailed differences. Consequently, a single-granularity approach struggles to ensure both semantic accuracy and robustness in cross-modal matching.

To enhance retrieval accuracy and flexibility, we propose a hybrid granularity retrieval strategy.It incorporates both coarse-grained and fine-grained retrieval granularity.Fine-grained retrieval operates at the token level, enabling precise identification and retrieval of subtle yet significant semantic features across different modalities, thereby improving accuracy. In contrast, coarse-grained retrieval combines adjacent tokens with similar semantics into larger units to capture local semantic consistency. This approach not only reduces noise from excessive refinement during the retrieval process but also enhances robustness in cross-modal retrieval.

To accurately capture clusters that convey consistent semantic information, we adopt a graph-based connected component clustering algorithm. This algorithm first constructs a semantic similarity matrix \( S \in \mathbb{R}^{n \times n} \), where each entry \( S_{ij} \) represents the similarity between two visual tokens \( v_i \) and \( v_j \).

We represent the visual tokens and their pairwise similarities as an undirected graph \( G = (V, E) \), where each node \( v_i \in V \) represents a visual token, and an edge \( e_{ij} \in E \) exists between nodes \( v_i \) and \( v_j \) if \( S_{ij} \geq \epsilon \) for a similarity threshold \( \epsilon \). This thresholding forms edges only between tokens with strong semantic connections, making the graph sparse and more manageable.

To identify clusters of semantically consistent tokens, we perform connected component clustering on \( G \). Each connected component \( C_k \subseteq V \) represents a group of tokens that are highly similar to each other. Thus, each cluster \( C_k = \{v_i \mid v_i \in C_k\} \) can be considered a coarse-grained semantic unit.

After identifying the connected components, we proceed to pool the tokens within each component. For each coarse-grained unit \( C_k \), we compute an average vector:

\begin{equation}
    \mathbf{c}_k = \frac{1}{|C_k|} \sum_{v_i \in C_k} v_i
\end{equation}

which serves as a representative vector for the cluster. This approach reduces the number of tokens involved in the final retrieval process, enhancing retrieval efficiency while maintaining semantic consistency.

After retrieval, each coarse-grained unit \( C_k \) is reverted to its original vector for subsequent processing. This algorithm not only significantly reduces computational costs during retrieval but also enhances the capability to capture multimodal semantic consistency, thereby improving the overall retrieval efficiency and performance of the model.

\subsubsection{Post-Retrieval:Token Router}

In previous sections, we successfully extracted core visual tokens that significantly influence the final generation results. However, we consider that there is still a considerable amount of semantic redundancy among these tokens.

As discussed in Sec \ref{sec:3.2}, both the external redundancy and internal redundancy of visual representations can burden the model's generation process.
Compared to text data, visual data often contains multiple tokens conveying similar semantics. This internal semantic redundancy can substantially increase the computational burden during the generation process. Especially in semantic retrieval, tokens with similar semantics are more likely to be retrieved simultaneously, leading to a large number of duplicate tokens that reduce retrieval efficiency in coarse-grained retrieval tasks.

Consequently, the next crucial step is to eliminate redundancy among highly similar tokens to avoid the repeated transmission of the same or similar semantic information, thereby further enhancing the model's computational efficiency and the clarity of its information representation. This process can be seen as a refinement of the retrieval results from the previous stage, particularly focusing on filtering redundant tokens within coarse-grained clusters, preserving key information while eliminating unnecessary repetitions.

Specifically, we develop a token routing mechanism based on the retrieved sequence and leverage a semantic token router \( \mathcal{R} \) to identify tokens that closely align with the query. Let \( \mathbf{T} = \{t_1, t_2, \dots, t_n\} \) represent the set of retrieved tokens, each associated with a similarity score \( s_i \) relative to the query. We define a routing threshold \( \tau \) such that only tokens satisfying \( s_i \geq \tau \) are retained.

To ensure semantic uniqueness among the retained tokens, we introduce a redundancy filter. For any two tokens \( t_i \) and \( t_j \) that satisfy \( \text{similarity}(t_i, t_j) \geq \delta \), where \( \delta \) is a predefined redundancy threshold, we retain only the token with the higher similarity score to the query. This filtering process can be formalized as follows:
\begin{equation}
\mathbf{T}_{\text{filtered}} = \{ t_i \mid s_i \geq \tau \text{ and } \forall t_j, \text{similarity}(t_i, t_j) < \delta\}.
\end{equation}

By enforcing both the route threshold \( \tau \) and redundancy threshold \( \delta \), we ensure that each retained token \( t_i \in \mathbf{T}_{\text{filtered}} \) is semantically unique, effectively preventing the redundant transmission of information by similar tokens. This mechanism achieves an optimal trade-off between retrieval efficiency and performance.

% This token routing mechanism not only enhances the model's computational efficiency but also improves the accuracy of semantic matching, leading to a more concise and effective final retrieval output. Through this approach, we ensure that the model transmits richer semantic information using fewer tokens during the generation process, ultimately enhancing the overall performance of the multimodal model.

\section{Experiments}

\begin{figure}[t]
    \hfill % 将图像推向右边
     % \hspace{-1em} % 调整左侧位置
     \includegraphics[width=1.0\linewidth]{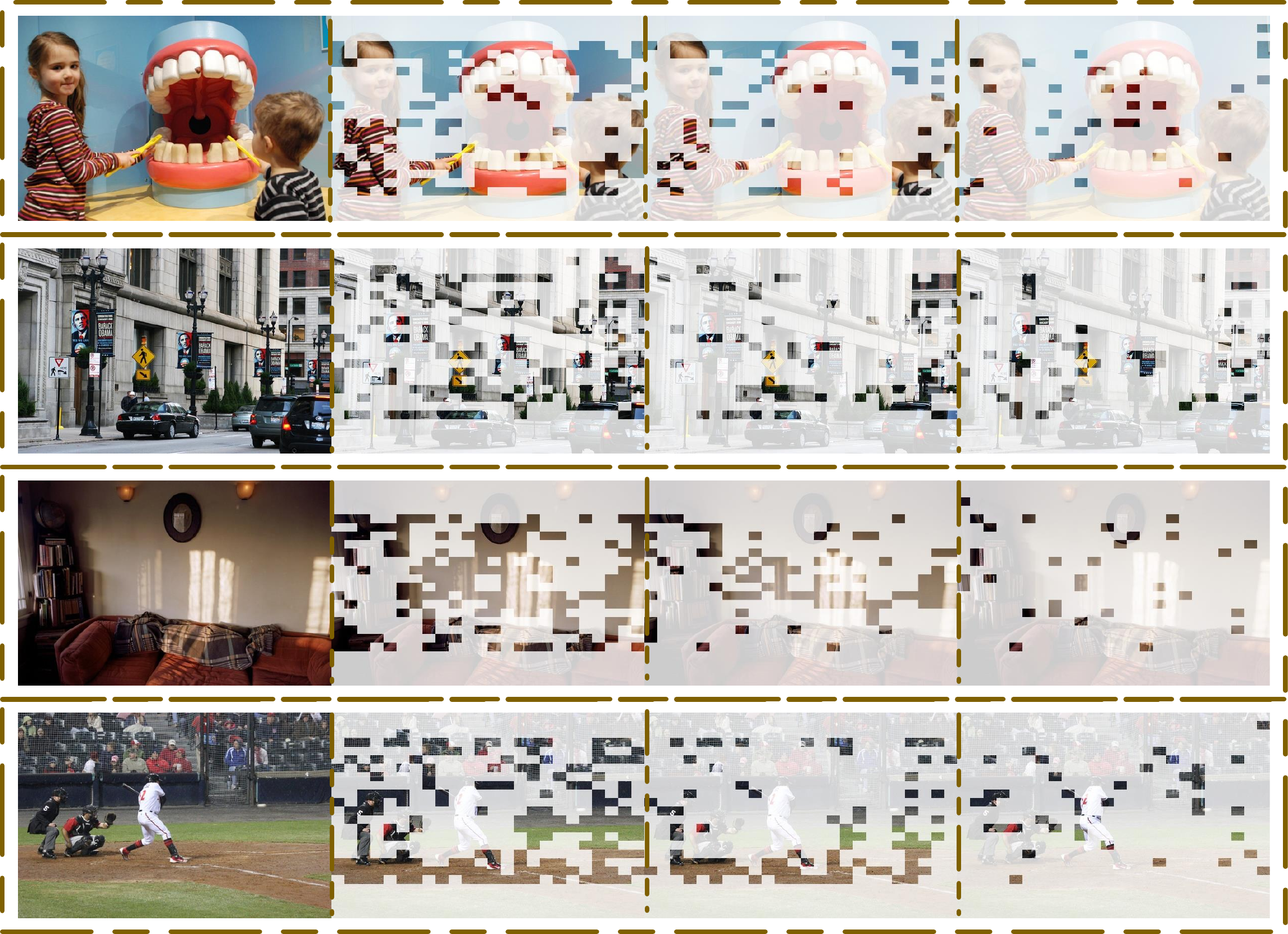} % 适当调整宽度
      % \vspace{-18em} % 减小标题与图片之间的距离
    \caption{
\textbf{Visualization of PAR}.From left to right, we change the ratio of retrieval and the visual tokens become increasingly sparse. In the utmost right is the final result of PAR.
}
    \label{fig:visual}
\end{figure}

\subsection{Experimental Setting}

\textbf{Datasets}.To accurately assess the practical effectiveness of our method, we selected four types of popular visual question-answering benchmarks, including (1)General visual question answering benchmark: GQA\cite{Hudson2019}
(2)Hallucination benchmark: POPE\cite{li2023}
(3)Optical character-based visual question answering benchmark: TextVQA\cite{singh2019}
(4)Comprehensive benchmark:MME\cite{fu2024},MMBench\cite{liu2024}, MMVet\cite{yu2023}.

\textbf{Model}.We first applied our method to LLaVA-1.5\cite{liu2023}, an open-sourced multimodal large model designed for tasks such as visual question answering and image captioning.LLaVA-1.5 uses CLIP\cite{radford2021} as the visual encoder and a LLaMA-based \cite{touvron2023}backbone large language model. The encoder and model are connected with a linear projector. To ensure semantic consistency, we use the CLIP text encoder as a text feature extractor in our method.

Notably, our approach does not require additional training or fine-tuning, setting it apart from most mainstream methods. All experiments were conducted on an NVIDIA A100 80G GPU.

\subsection{Performance evaluation}

In Table \ref{main}, we present PAR's performance on several visual question-answering benchmarks.

Compared to the baseline LLaVA-1.5 model, our approach maintains approximately 97\% of the original model's accuracy across diverse datasets, using only around 11\% of the visual tokens. This yields a tenfold increase in efficiency. Notably, on the multimodal hallucination benchmark POPE, PAR surpasses the original model in accuracy, underscoring its effectiveness in handling hallucination issues.

To further validate PAR’s efficiency, we benchmarked it against two state-of-the-art methods: LLaVA-PruMerge \cite{shang2024} and Fast-V \cite{chen2025image}. Using identical experimental conditions and reproduced open-source code and weights, PAR consistently outperformed these methods both in terms of accuracy and token pruning efficiency.

These results highlight that our method, by effectively reducing redundancy in visual tokens, achieves near-baseline performance with a fraction of the computational cost. Furthermore, its strong results on hallucination benchmarks illustrate PAR’s capability to minimize irrelevant contextual interference, demonstrating its robustness and precision in multimodal inference.

\begin{table*}[t]
\centering
 \renewcommand{\arraystretch}{1.5}
\resizebox{1.0\linewidth}{!}{
\begin{tabular}{lcccccccc}
\toprule
\textbf{Method}       & \textbf{GQA}       & \textbf{VQA$^{\text{text}}$} & \textbf{POPE}   & \textbf{MME$_{\text{per}}$} & \textbf{MME$_{\text{cog}}$} & \textbf{MMBench} & \textbf{MMvet} & \textbf{Avg} \\ 
\midrule
LLaVA-v1.5-7B         & 61.23 / 100        & 60.12 / 100       & 85.3 / 100      & 1529.77 / 100    & 439.29 / 100     & 65.42 / 100      & 29.3 / 100     & \textbf{100/100} \\ 
LLaVA-PruMerge        & 54.41 / 19.82      & 56.94 / 20.49     & 82.33 / 19.96   & 1501.84 / 20.24  & 435.64 / 20.24   & 65.70 / 20.79    & 28.8 / 20.5    & 96.43 / 20.29 \\ 
Fast-V                & 48.04 / 20.00      & 45.00 / 20.00     & 77.48 / 20.00   & 1424.70 / 20.00  & 405.00 / 20.00   & 60.64 / 20.00    & 24.3 / 20.00   & 86.29 / 20 \\ 
\rowcolor{pink}
\textbf{PAR (Ours)}   & 57.38 / 12.56      & 57.06 / 12.65     & 87.18 / 12.54   & 1527.71 / 10.57  & 432.86 / 10.57   & 64.33 / 8.95     & 27.2 / 13.44   & 97.03($\downarrow$\textcolor{red}{2.97}) / 11.60($\downarrow$\textcolor{blue}{88.4}) \\ 
\bottomrule
\end{tabular}
}
\caption
{\textbf{Model performance comparison across various datasets.}
Each metric is shown as "Accuracy / Token ratio". Compared to the baseline, our method results in a 2.97\% decrease in accuracy, but achieves an 88.4\% reduction in token usage, striking a better balance between performance and efficiency.}
\label{main}
\vspace{-1mm}
\end{table*}

\subsection{Ablation Study}

This section establishes the purpose and structure of the ablation study clearly, setting up an in-depth examination of each component's contributions.

\begin{table*}[htb]
\centering
 \renewcommand{\arraystretch}{1.2}
    \begin{subtable}{.33\textwidth}
        \begin{tabular}{lcc}
\textbf{Distance Metric}     & \textbf{POPE}   & \textbf{MMB}                                       \\ 
\midrule
L1      & 84.18     & 64.24        \\  
L2   & 82.01      & 63.31                      \\        
Lp    & 68.89        & 56.94              \\
Linf   & \colorbox[rgb]{0.89, 0.89, 0.89}{87.18}       & \colorbox[rgb]{0.89, 0.89, 0.89}{64.33}        \\
Inner Product  & 83.20      & 62.07 \\
\end{tabular}
        \subcaption{\textbf{Distance Metric.} }
        \label{tab:abalation_distance metric}
    \end{subtable}
    \begin{subtable}{.33\textwidth}
        \centering
        \begin{tabular}{ccc}
\textbf{Query Condition}    & \textbf{POPE}   & \textbf{MMB}     \\
\midrule
Original Prompt   & 82.38   &  61.29 \\
Key word Prompt  & 84.25   & 63.57 \\
Prompt Rewriting   & \colorbox[rgb]{0.89, 0.89, 0.89}{87.18} & \colorbox[rgb]{0.89, 0.89, 0.89}{64.33}  \\
     &  \\
      & \\
\end{tabular}
        \subcaption{\textbf{Query Prompt.}}
        \label{tab:abalation_query}
    \end{subtable}
        \begin{subtable}{.33\textwidth}
        \centering
        \begin{tabular}{ccc}
\textbf{Token number}    & \textbf{POPE}   & \textbf{MMB}     \\
\midrule
1*1 Token   & 83.91   &  63.86 \\
2*2 Tokens   &84.34   & 62.88\\
3*3 Tokens   &82.91 & 62.76 \\
Semantic Tokens   &\colorbox[rgb]{0.89, 0.89, 0.89}{87.18}  &\colorbox[rgb]{0.89, 0.89, 0.89}{64.33} \\
      & \\
\end{tabular}
        \subcaption{\textbf{Retrieval Granularity.} }
        \label{tab:abalation_retrieval}
    \end{subtable}%
\caption{\textbf{PAR ablation experiments} on POPE and MMB.The selected configurations are marked in \colorbox[rgb]{0.89, 0.89, 0.89}{gray}.}
\label{tab:ablation}
\end{table*}

\begin{figure}[t]
    \hfill % 将图像推向右边
     \hspace{-1em} % 调整左侧位置
     \includegraphics[width=1.0\linewidth]{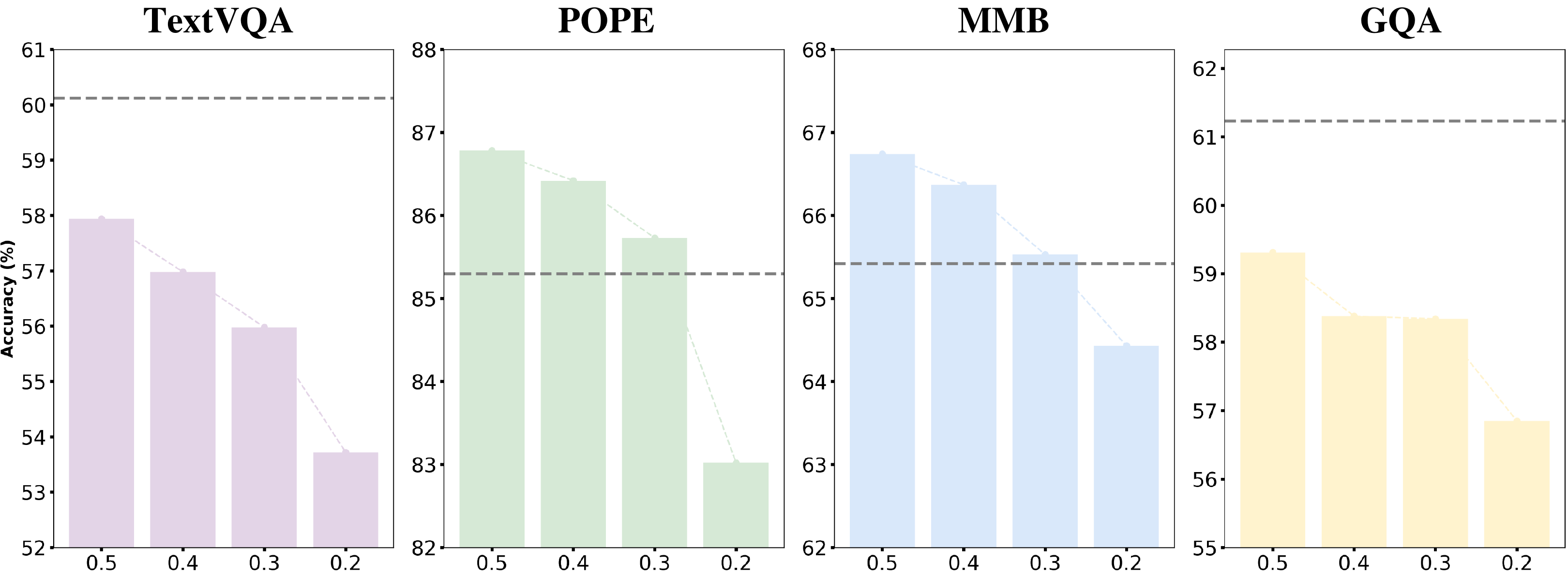} % 适当调整宽度
      % \vspace{-18em} % 减小标题与图片之间的距离
    \caption{
\textbf{Accuracy results} on four datasets (Text VQA, POPE, MMbench, GQA) using only the \textbf{direct retrieval} with token ratio of 50\%, 40\%, 30\%, and 20\%.
}
\label{fig:direct_retrieval}
\end{figure}

\begin{figure}[t]
    \hfill % 将图像推向右边
     \hspace{-1em} % 调整左侧位置
     \includegraphics[width=1.0\linewidth]{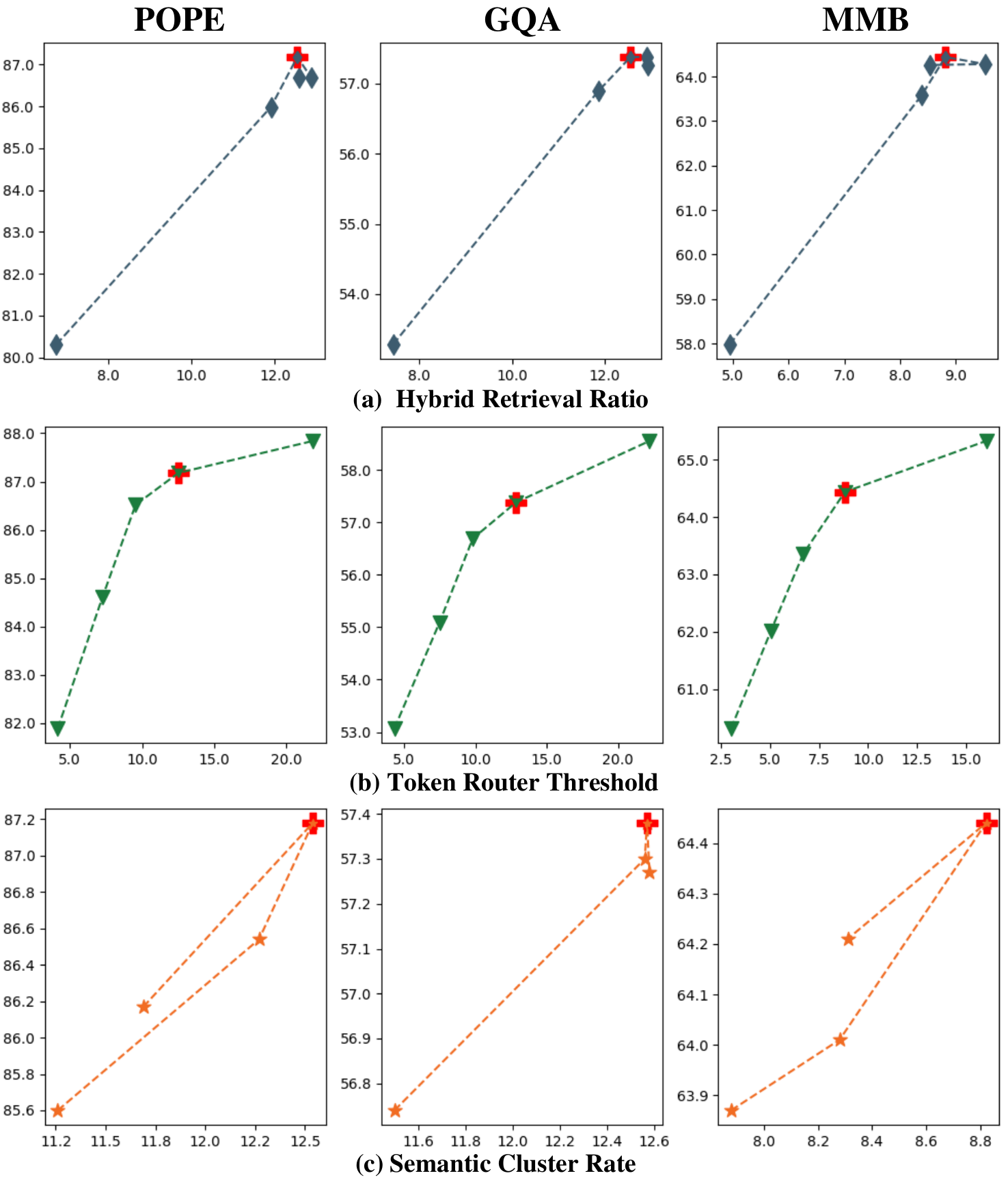} % 适当调整宽度
      % \vspace{-18em} % 减小标题与图片之间的距离
    \caption{
\textbf{Hyperparameters ablation results} about \textbf{Hybrid Retrieval Ratio} ,\textbf{Token Router Threshold} and 
\textbf{Semantic Cluster Rate} across three datasets.
To illustrate the trade-off between performance and efficiency, we use Token Ratio as the x-axis and Accuracy as the y-axis. The red sign represents the selected parameters.
}
\label{fig:abl}
\end{figure}

% For retrieval-based systems, the choice of retrieval algorithm directly impacts overall efficiency and inference performance. Different retrieval algorithms vary in terms of processing speed, memory overhead, and semantic matching accuracy, making it critical to select the appropriate algorithm.

% In this experiment, we conducted an ablation study on the retrieval algorithms used in TRSM  to evaluate their effect on final accuracy. By testing various retrieval algorithms under the same conditions, we quantified their differences in processing efficiency and accuracy, allowing for a deeper analysis of their specific impact on model performance. The results, as shown in Table 3, highlight the relative performance of each retrieval algorithm during the model inference process.

\textbf{Retrieval Ratio.} 
Figures \ref{fig:direct_retrieval} and \ref{fig:abl} illustrate how different retrieval retention ratios impact the final results. We examined the effects of varying retrieval ratios in both direct retrieval and hybrid retrieval strategies.

For \textbf{direct retrieval}, we removed all the proposed components and relied solely on the most basic semantic retrieval algorithm for token pruning. 
Here, the number of retrieved tokens corresponds directly to the final count of tokens involved in generation. As the retrieval ratio decreased, the number of tokens used in generation gradually reduced, leading to a decline in the model’s accuracy on the dataset.
Notably, for the POPE and MMB datasets, the accuracy only dropped below the original model’s baseline when the retrieval ratio was reduced below 30\%. This demonstrates the effectiveness of our approach in mitigating hallucination phenomena in large models.

For \textbf{ hybrid retrieval},  due to the inherent randomness in image semantics, predicting the exact number of retained tokens is challenging. Thus, we used the initial retrieval count as a reference standard and tested five retrieval ratios—20\%, 50\%, 60\%, 75\%, and 100\%. 
The experimental results on the three datasets showed that the trends for the first three ratios were similar to direct retrieval, with accuracy gradually improving as the retrieval amount increased. However, when the retrieval ratio reached 60\% , further increasing the retrieval ratio caused a drop in accuracy. We hypothesize that the effective data regions of the majority of images are concentrated within the 60\% retrieval range. Increasing the retrieval amount beyond this threshold may lead the model into hallucination phenomena, resulting in a decrease in accuracy.

Moreover, the results from hybrid retrieval reveal the proportional relationship between external and internal redundancy. When the initial retrieval ratio is set to 60\% and the final optimized ratio is reduced to 10\%, the model maintains a similar accuracy to the original. This suggests that the ratio of effective information to external and internal redundancy is approximately 1:4:5.
This insight provides a valuable perspective for further exploring redundancy patterns in multimodal reasoning, facilitating more precise optimization of token selection strategies.

\textbf{Token Router Threshold.}
Figure \ref{fig:abl} illustrates the effect of varying routing thresholds on experimental results. The Token Router is designed to reduce internal redundancy among visual tokens, particularly by distinguishing semantically redundant tokens produced during semantic clustering, thereby lowering the final token count.

The routing threshold directly influences the number of retained tokens, and we tested five threshold values (0.4, 0.5, 0.55, 0.6, and 0.7) to assess their impact on performance. Since the router primarily eliminates tokens contributing redundant semantic information, changes in threshold values generally align with the growth trend seen in direct retrieval: as more tokens are retained, accuracy improves. However, results show that at a threshold of 0.7, the rate of accuracy improvement declines notably as the retrieval count increases. An excess of tokens at this point reduces the method’s efficiency, disrupting the balance between performance and efficiency. Based on these results and practical considerations, we chose 0.6 as the optimal router threshold.

\textbf{Clustering Rate.}
Figure \ref{fig:abl} demonstrates the effect of varying semantic clustering rates on experimental results. Selecting an appropriate clustering rate is crucial for optimizing retrieval efficiency. If the rate is set too high, it may fail to effectively aggregate nearby semantic information, while a rate set too low can introduce noise, diminishing clustering effectiveness. To assess how different clustering rates influence generation accuracy and token retention, we tested four rates: 0.6, 0.7, 0.8, and 0.9. The results show that for the first three rates, our method follows a pattern similar to direct retrieval, with accuracy increasing as more tokens are retained. Based on these findings, we selected 0.8 as the optimal semantic clustering rate.

\textbf{Distance Metric.} 
Table \ref{tab:abalation_distance metric} presents our investigation into different distance-based retrieval algorithms.
In retrieval tasks, selecting an appropriate distance function is critical for effective similarity measurement. Different distance functions can have a significant impact on similarity computation, thereby affecting the final retrieval results and the model's performance.
We evaluated five different distance functions: L1 distance, L2 distance, Lp distance, Linf distance, and Inner product. Among these, Linf distance—highlighting the maximum individual difference—yielded the best results, demonstrating its suitability for tasks that require sensitivity to distinct feature variations.

\textbf{Query Condition.}
Table \ref{tab:abalation_query} investigates the impact of different query structures on retrieval performance. By comparing the original prompt, keyword extraction, and template-based rewriting, we explore how query optimization can enhance retrieval. The original prompt serves as a baseline, while keyword extraction improves semantic clarity by eliminating redundancy. Template-based rewriting further aligns text and visual embeddings, boosting retrieval accuracy. Experimental results show that the predefined query rewriting framework significantly improves performance."

\textbf{Retrieval Granularity.} 
In table \ref{tab:abalation_retrieval} we investigated the impact of four different retrieval approaches on the experimental outcomes.We first employ fine-grained retrieval under different settings, and later explore a hybrid retrieval strategy that combines both coarse-grained and fine-grained approaches.
For fine-grained retrieval, we used token blocks of sizes 1×1, 2×2, and 3×3 as the basic retrieval units. In contrast, for hybrid retrieval, we employed semantic clusters based on semantic clustering as the retrieval units. Experimental results show that the hybrid retrieval based on semantic clustering effectively incorporate surrounding semantic information, thereby achieving the highest benchmark accuracy.

\subsection{Efficiency Analysis}

\begin{table}[t] \Huge
\centering
\renewcommand{\arraystretch}{1.5}
\resizebox{1.0\linewidth}{!}{
\begin{tabular}{lccccc}
\toprule
\Huge \textbf{Method} & \Huge \textbf{FLOPs} & \Huge \textbf{Total Memory} & \Huge \textbf{Prefill Time} & \Huge \textbf{Activation} & \Huge \textbf{KV Cache} \\ 
& \Huge \textbf{(TB)} & \Huge \textbf{(GB)} & \Huge \textbf{(ms)} & \Huge \textbf{(GB)} & \Huge \textbf{(MB)} \\ 
\midrule
 LLaVA          &  8.2                  & 21.8                 & 59.1                & 3.9                  & 323 \\ 
PruMerge      & 2                    & 14.8                 & 19.3                & 0.68                 & 81.8 \\ 
\rowcolor{pink} 
PAR                  & \textbf{1.4}  & \textbf{14.2} & \textbf{18.5}  & \textbf{0.43}  & \textbf{54.5}  \\ 
& $(\downarrow\textcolor{blue}{6.8})$  & $(\downarrow\textcolor{blue}{7.6})$ 
& $(\downarrow\textcolor{blue}{40.6})$ & $(\downarrow\textcolor{blue}{3.47})$
& $(\downarrow\textcolor{blue}{268.5})$ \\
\bottomrule
\end{tabular}
}
\caption{\textbf{Efficiency Analysis}. We use LLaVA-v1.5-7B as a baseline, the precision is fp16 and batchsize=1. All the data are estimated using a theoretical model.}
\label{tab:eff}
\vspace{-1mm}
\end{table}

To efficiently evaluate the computational performance of our method, we conducted a theoretical analysis of factors such as latency and memory usage using the Roofline tool based on LLMviewer\cite{yuan2024}.
Using the LLaVA-1.5 7B model as an example, we analyzed the multimodal large model inference process in typical scenarios. This model processes images with a resolution of 336×336 pixels, which are converted into 576 visual tokens through the CLIP model, combined with a prompt input of approximately 40 prompt tokens.

LLaVA-PruMerge\cite{shang2024} achieved a compression ratio of about 20 \%, reducing the visual tokens to 116. In contrast, our method, while maintaining similar accuracy on VQA tasks, achieved an 11\% compression ratio, reducing the visual tokens to around 64.

As shown in Table \ref{tab:eff}, our method significantly improved model inference speed and reduced memory consumption. Specifically, for generating the first token, prefill time was reduced to 31.3\% of the original, and the activation during inference was reduced to 11\%. This makes our method more suitable for deploying large models in resource-constrained environments.

\section{Conclusions}
% \setstretch{1.25}

In this paper, we introduced PAR(Prompt-Aware Token Reduction), an efficient approach for reducing visual tokens in multimodal language models. Inspired by human visual processing and the observed redundancy in visual contexts, PAR adaptively identifies and clusters relevant visual tokens through semantic retrieval. Then followed by a token routing mechanism to refine and retain only the most significant tokens. 
Experiments demonstrated that PAR achieves a better balance between performance and efficiency than previous works. It achieves an 83\% reduction in FLOPs, decreases prefill time to 31.3\% of the original, and attains a compression ratio of 89\%, all while maintaining 97\% of the baseline accuracy across a wide range of visual question-answering and reasoning tasks. We hope our work inspires further exploration into the interplay between efficiency and performance in MLMMs.
{
    \small
    % \nocite{*}
    \bibliographystyle{ieeenat_fullname}
    \bibliography{main}
}


\end{document}